\documentclass[10pt,twocolumn,letterpaper]{article}

\usepackage{cvpr}
\usepackage{times}
\usepackage{epsfig}
\usepackage{graphicx}
\usepackage{amsmath}
\usepackage{amssymb}
\usepackage{multirow}
\usepackage{bm}
\usepackage{array}

\usepackage[breaklinks=true,bookmarks=false]{hyperref}

\cvprfinalcopy 

\def\cvprPaperID{****} 

\begin{document}

\title{Attacks on State-of-the-Art Face Recognition using Attentional Adversarial Attack Generative Network}

\author{Qing Song, Yingqi Wu and Lu Yang\\
Pattern Recognition and Intelligent Vision Lab, Beijing University of Posts and Telecommunications\\
{\tt\small \{priv, wuyqq, soeaver\}@bupt.edu.cn}
}

\maketitle

\begin{abstract}
With the broad use of face recognition, its weakness gradually emerges that it is able to be attacked. So, it is important to study how face recognition networks are subject to attacks. In this paper, we focus on a novel way to do attacks against face recognition network that misleads the network to identify someone as the target person not misclassify inconspicuously. Simultaneously, for this purpose, we introduce a specific attentional adversarial attack generative network ($A^{3}GN$) to generate fake face images. For capturing the semantic information of the target person, this work adds a conditional variational autoencoder and attention modules to learn the instance-level correspondences between faces. Unlike traditional two-player GAN, this work introduces face recognition networks as the third player to participate in the competition between generator and discriminator which allows the attacker to impersonate the target person better. The generated faces which are hard to arouse the notice of onlookers can evade recognition by state-of-the-art networks and most of them are recognized as the target person.
\end{abstract}

\section{Introduction}

Neural network is widely used in different tasks in society which is profoundly changing our life. Good algorithm, adequate training data, and computing power make neural network supersede human in many tasks, such as face recognition. Face recognition can be used to determine which one the face images belong to or whether the two face images belong to the same one. Applications based on this technology are gradually adopted in some important tasks, such as identity authentication in a railway station and for payment. Unfortunately, it has been shown that face recognition network can be deceived inconspicuously by mildly changing inputs maliciously. The changed inputs are named adversarial examples which implement adversarial attacks on networks. 

Szegedy {\it et al.}~\cite{Intrigue} present that adversarial attacks can be implemented by applying an imperceptible perturbation which is hard to be observed for human eyes for the first time. Following the work of Szegedy, many works focus on how to craft adversarial examples to attack neural networks. Neural network is gradually under suspicion. The works on adversarial attacks can promote the development of neural network. Akhtar {\it et al.}~\cite{ThreatSurvey} review these works' contributions in the real-world scenarios. Illuminated by predecessor's works, we also do some research about adversarial attack.  


\begin{figure}[t]
\begin{center}
\includegraphics[width=1.0\linewidth]{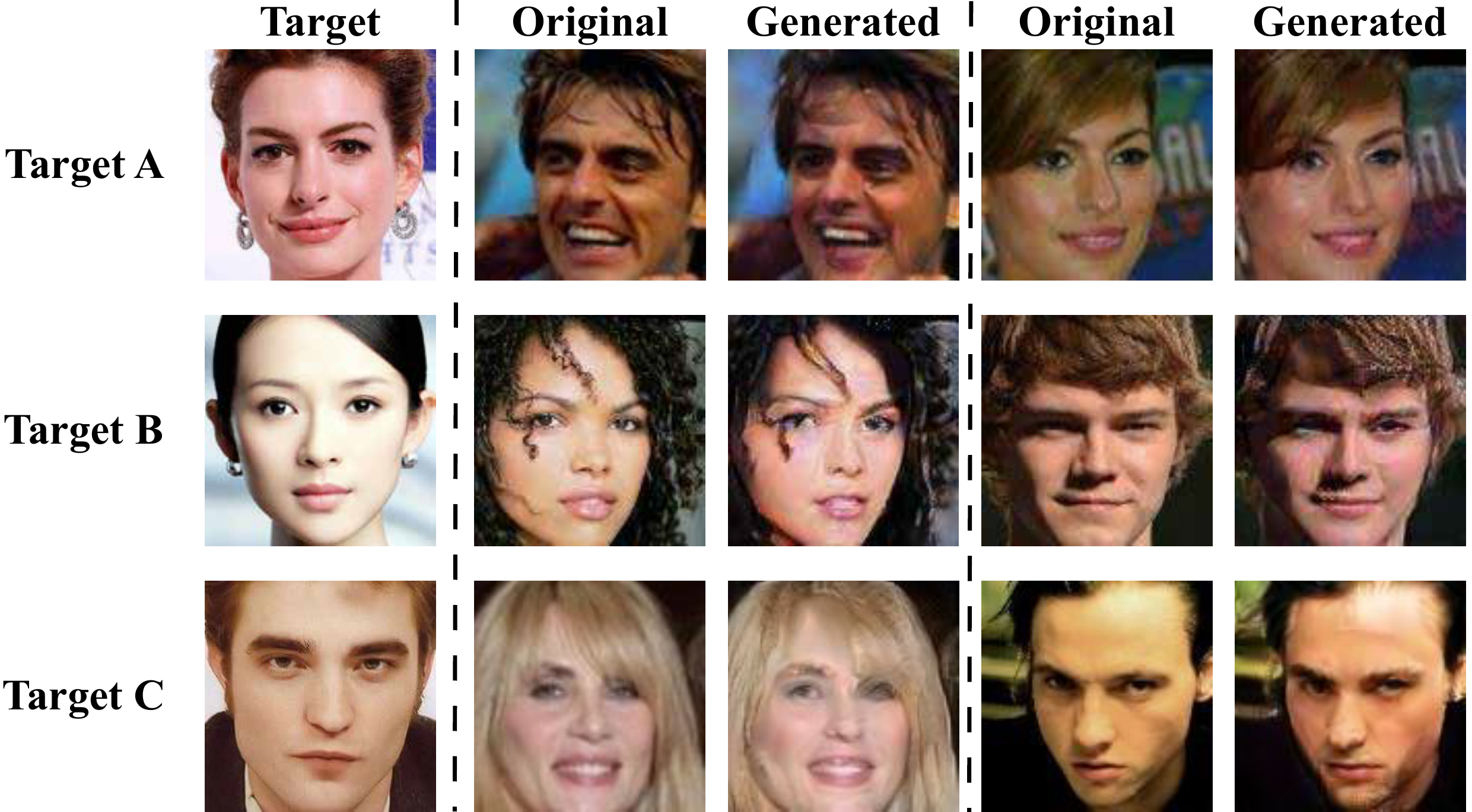}
\end{center}
\caption{Adversarial attack results in our work. The first column is the target face. The 2nd and 4th columns are the original images and the rest are the generated images. Given target images, our work is to generate images similar to the original faces but classified as the target person.}
\label{fig:description}
\end{figure}
Most of adversarial attacks aim at misleading classifier to a false label, not a determined specific label. Besides, attacks on image classifier can not be against face recognition networks. Existing works produce perturbation on the images~\cite{gradient, Intrigue, perturbation}, do some makeup to faces and add eyeglass, hat or occlusions~\cite{eyeglass1, eyeglass2, occlusion} to faces. And their adversarial examples are fixed by the algorithms which are not flexible for attacks. These algorithms can not accept any images as inputs. Our goal is to generate face images which are similar to the original images but can be classified as the target person shown in Fig.~\ref{fig:description}. The method manipulating the intensity of input images directly is intensity-based. Our work uses geometry-based method to generate adversarial examples. In our work, we use generative adversarial net (GAN) ~\cite{GAN} to produce adversarial examples which are not limited by data, algorithms or target networks. It can accept any faces as inputs and convert them to adversarial examples for attacks. To generate adversarial examples, we present $A^{3}GN$ to produce the fake image whose appearance is similar to the origin but is able to be classified as the target person.

In face verification domain, whether the two faces belong to one person is based on the cosine distance between feature map in the last layer not based on the probability for each category. So $A^{3}GN$ pays more attention to the exploration of feature distribution for faces. To get the instance information, we introduce a conditional variational autoencoder to get the latent code from the target face, and meanwhile, attentional modules are provided to capture more feature representation and facial dependencies of the target face. For adversarial examples, $A^{3}GN$ adopts two discriminators -- one for estimating whether the generated faces are real called {\it normal discriminator}, another for estimating whether the generated faces can be classified as the target person called {\it instance discriminator}. Meanwhile, cosine loss is introduced to promise that the fake images can be classified as the target person by the target model. Our main contributions can be summarized into three-fold:

\begin{itemize}
\vspace{-0.1cm}
\item
We focus on a novel way of attacking against state-of-the-art face recognition networks. They will be misled to identify someone as the target person not misclassify inconspicuously in face verification according to the feature map not the probability. \vspace{-0.1cm}
\item
GAN is introduced to generate the adversarial examples different from traditional intensity-based attacks. Meanwhile, this work presents a new GAN named $A^{3}GN$ to generate adversarial examples which are similar to the origins but have the same feature representation as the target face.\vspace{-0.1cm}
\item
Good performance of $A^{3}GN$ can be shown by a set of evaluation criteria in physical likeness, similarity score, and accuracy of recognition.
\end{itemize}
\section{Related work}
\subsection{Face recognition}
We witness the great development and success of convolutional neural network in face recognition so far. With the development of advanced architectures and discriminative learning approaches, face recognition performance has been boosted to an unprecedented level. Face recognition can be categorized as face verification and face identification. In our work, we focus on face verification which determines whether a pair of faces belong to the same person and the latter classifies a face to a specific identity. Learning discriminative deep face representation through large-scale face identity classification was proposed by ~\cite{deepface1, deepface2, deepid, deepid2, deepid2+, deepid3}. More and more CNN-based approaches are absorbed in loss functions~\cite{sphereface, cosineface, arcface, centerloss, contrastive}. Our goal is to generate adversarial examples with $A^{3}GN$ to attack these state-of-the-art face recognition networks in face verification.
\subsection{Adversarial attack}
With the remarkable accuracy, neural network gets access to many important domains in society, such as self-driving cars, surveillance and identity authentication. The security problem of neural networks has become a critical problem. Szegedy {\it et al.}~\cite{Intrigue} reveal the perturbation which can fool DNN for the first time. Moosavi-Dezfooli {\it et al.}~\cite{Moosavi-Dezfooli} demonstrate that `universal perturbation' can fool the classifier by any image in most type of models. I.J.Goodfellow {\it et al.}~\cite{gradient} present that the intrinsic reason for adversarial attack is the linearity and high-dimensions of inputs. Su {\it et al.}~\cite{onepixel} present a method to generate one-pixel adversarial perturbations to attack models using differential evolution in an extremely specific scenario. In the face recognition domain, Bose {\it et al.}~\cite{attackfaceDetect} craft adversarial examples by solving constrained optimization so that face detector can not detect faces. Sharif {\it et al.}~\cite{eyeglass2} propose a method focusing on facial biometric systems which can be widely used in surveillance and access control. Many works are proposed to explore more imperceptible adversarial examples to attack neural networks efficiently~\cite{C.Kanbak, L.Engstrom, N.Carlini, R.Huang, S.-M.Moosavi-Dezfooli, T.Miyato}.

In this paper, we focus on generating quasi-imperceptible adversarial examples to do white-box and targeted attacks.
\subsection{Generative adversarial network}
Generative adversarial networks~\cite{GAN} have achieved great performance and impressive results in image generation~\cite{radford, denton}, style transfer~\cite{ulvanov, johnson, gatys}, image-to-image translation~\cite{isola, cyclegan, bicycle} and representation learning~\cite{radford, salimans, mathieu}. Most works utilize conditional variables such as attributes~\cite{attribute1, stargan}. CycleGAN~\cite{cyclegan} preserves key attributes between the input and the translated images by a cycle consistency loss which has received a good improvement in unpaired image-to-image translation. Conditional VAEs~\cite{sohn} have shown the good performance for image-to-image translation which learn a mapping from input to output image. In~\cite{bicycle}, cVAE-GAN and cLR-GAN are used to learn a low-dimensional latent code and then map from a high-dimensional input to a high-dimensional output. In our work, we use conditional variational autoencoder GAN to learn the feature representation of the target person for generating adversarial examples from any faces to attack face recognition networks. 
\section{Attentional Adversarial Attack Generative Network}
\subsection{Threat model}
In our work, the adversary aims at fooling a face recognition network to recognize someone to another one which belongs to impersonation(targeted) attack. To achieve this purpose, we consider a white-box attack with the access of face recognition networks which has knowledge of the outputs, parameters, architectures, training data of target networks. Simultaneously, we also consider a black-box attack to test our $A^{3}GN$ transferability. We adopt state-of-the-art face recognition networks as target models. Different networks learn different feature representations for face images which will prove that $A^{3}GN$ can be applied to attack different networks. Our adversarial goal is to generate adversarial images which are similar to the original images from LFW datasets but recognized as the target person by imitating feature representation of the target person from target models. For this goal, we present $A^{3}GN$ for generating adversarial examples.
\subsection{A$^{3}$GN}
\begin{figure*}
\begin{center}
\includegraphics[width=0.9\linewidth]{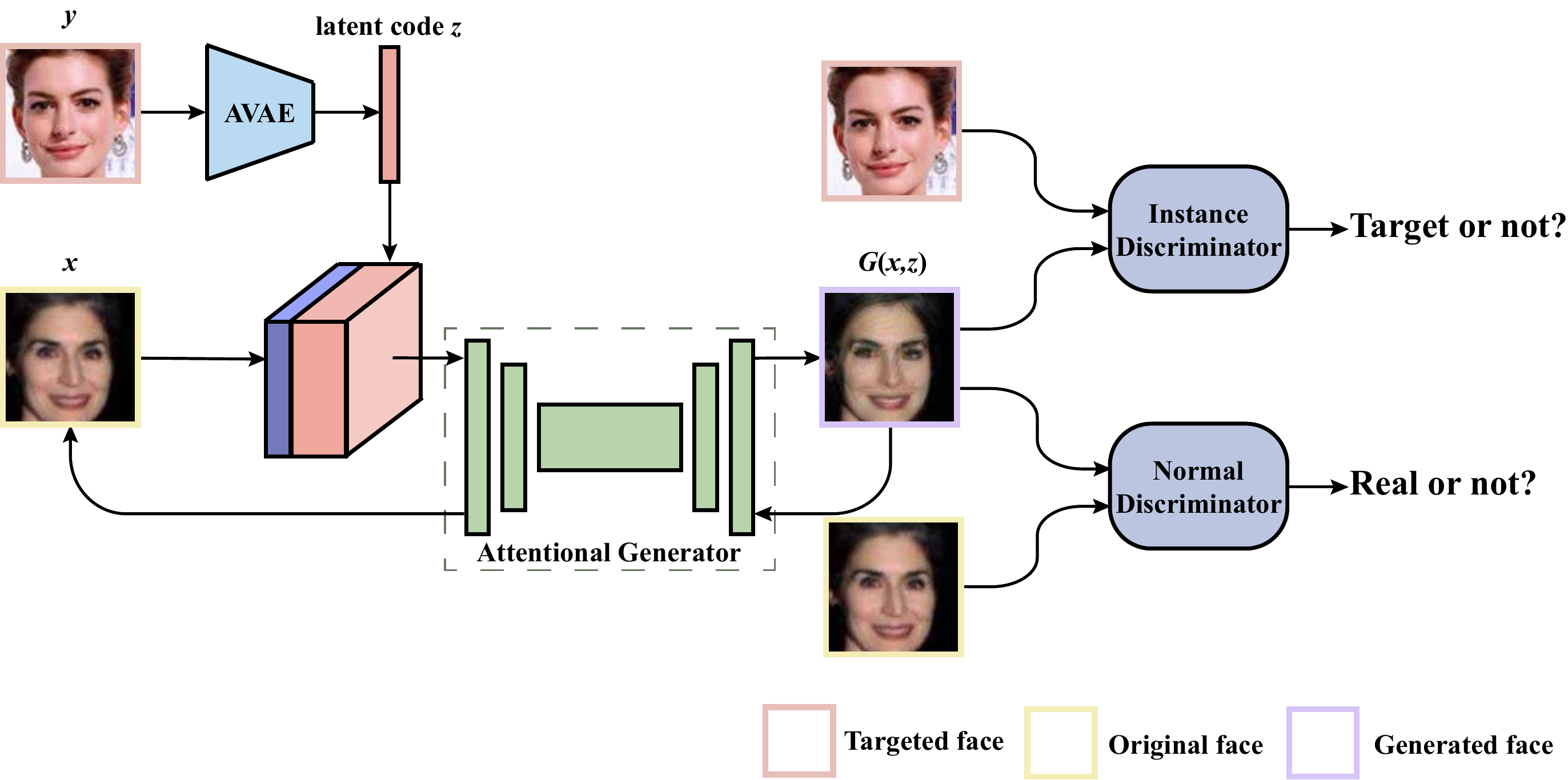}
\end{center}
\caption{Overview of $A^{3}GN$. {\it Attentional variational autoencoder (AVAE)} captures the latent code $z$ from target face $y$. And then the original face $x$ is concatenated with $z$ to generate $\hat{x}$, $G(x,z) \to \hat{x}$ in {\it attentional generator}. $G(x,z)$ is sent into {\it normal discriminator} to determine whether it is a real image or not with $x$ and sent into {\it instance discriminator} to determine whether it can be classified as the target person or not with $y$.}
\label{fig:a3gn}
\end{figure*}
We adopt conditional variational autoencoder generative adversarial networks which can learn targeted feature for generating adversarial examples to fool the target network. To guarantee that generated images are classified as the target person better, we introduce some attentional blocks into conditional variational autoencoder GAN to constitute $A^{3}GN$ shown in Fig.~\ref{fig:a3gn}.

\noindent\textbf{Conditional variational autoencoder GAN.} For exploring the feature distribution of different faces, we use cVAE-GAN to capture the instance information of different faces for the generator to produce the adversarial examples. Given a target image $y$, using an encoding function  $E$ learns a latent code $z$ of $y$, $E(y) \to z$. Generator $G_1$ combines $z$ and an input image $x$ to synthesize the output $\hat{x}$, $G_1(x,z) \to \hat{x}$. Normal discriminator $D_1$ determines whether $\hat{x}$ is real or not. To make the generated images indistinguishable from real images, we adopt an adversarial loss:
\begin{equation}\label{eq:adloss}
\mathcal{L}_{adv}=  \mathbb{E}_x[logD_1(x)] + \mathbb{E}_{x,z}[log(1-D_1(G_1(x,z)],
\end{equation}
where generated image $G_1(x,z)$ learns the latent code $z$ from the target image $y$, while normal discriminator $D_1$ tries to distinguish $G_1(x,z)$ between real and fake image. Normal discriminator $D_1$ tries to maximize $D_1(x)$ which is opposite to the generator $G_1$.

For the stability of training and high quality generated images, we replace Eq.~\ref{eq:adloss} with Wasserstein GAN objective with gradient penalty~\cite{wloss1}~\cite{wloss2}:
\begin{equation}\begin{split}\label{eq:wloss}
\mathcal{L}_{adv}=\mathbb{E}_x[logD_1(x)]-\mathbb{E}_{x,z}[D_1(G_1(x,z))]\\
-\lambda_{gp}\mathbb{E}_{x'}[(\left \|\nabla_{x'}D_1(x')\right \|_2-1)^2],
\end{split}\end{equation}
where $x'$ is sampled between a pair of a real and a generated images. And $\lambda_{gp}$ is set to 10.

To preserve the content of the input images while changing instance-level information and a part of feature representation of the inputs, we introduce a cycle-consistency loss~\cite{cyclegan} to the generator as reconstruction loss:
\begin{equation}\label{eq:recloss}
\mathcal{L}_{rec}=  \mathbb{E}_{x,z}[\lVert x-G_2(G_1(x,z))\rVert_1],
\end{equation}
where $G_2$ is used to take in the generated image $G_1(x,z)$ as input and reconstruct the original image $x$. The reconstruct loss adopts the $\ell_1$ norm. Here, $G_1$ and $G_2$ are two different generators with inputs of different dimensions.
 
\noindent\textbf{Instance discriminator.} In this work, we propose an instance discriminator $D_2$ as third-player to participate in the competition which brings about impersonating target faces better. For generating images with the similar feature representation to the target image, we adopt face recognition network as the instance discriminator directly. For a given input image $x$, and a latent code $z$ from the target image $y$, $E(y) \to z$, our goal is to translate $x$ into $\hat{x}$, $G_1(x,z) \to \hat{x}$, which can be classified as $y$ by $D_2$. To achieve this condition, we adopt a cosine loss, defined as:
\begin{equation}\begin{split}\label{eq:cosloss}
\mathcal{L}_{cos}=1-SIM(y,G_1(x,z))=1-\cos\theta\\
=1-\mathbb{E}_{x,y,z}[\cfrac{D_2(y) \cdot D_2(G_1(x,z))}{\left \| D_2(y) \right \|\cdot\left \| D_2(G_1(x,z)) \right \|}],
\end{split}
\end{equation}
where $D_2$ is the instance discriminator, and $D_2(y)$ and $D_2(G_1(x,z))$ mean the feature representation of $y$ and $G_1(x,z)$. Minimizing cosine loss can minimize the difference between generated image $G_1(x,z)$ and target image $y$ in space which brings benefit to generating adversarial examples. The objective functions are defined as,
\begin{equation}\label{eq:dloss}
\mathcal{L}_{D_1}=-\mathcal{L}_{adv},
\end{equation}
\begin{equation}\label{eq:gloss}
\mathcal{L}_{G}=\mathcal{L}_{adv}+\lambda_{rec}\mathcal{L}_{rec}+\lambda_{cos}\mathcal{L}_{cos},
\end{equation}
where $\lambda_{rec}$ and $\lambda_{cos}$ are hyper-parameters that control the relative importance of reconstruction loss and cosine loss respectively compared to the adversarial loss. In our work, we use $\lambda_{rec}$ = 10 and $\lambda_{cos}$ = 10.
\begin{figure}[t]
\begin{center}
\includegraphics[width=0.95\linewidth]{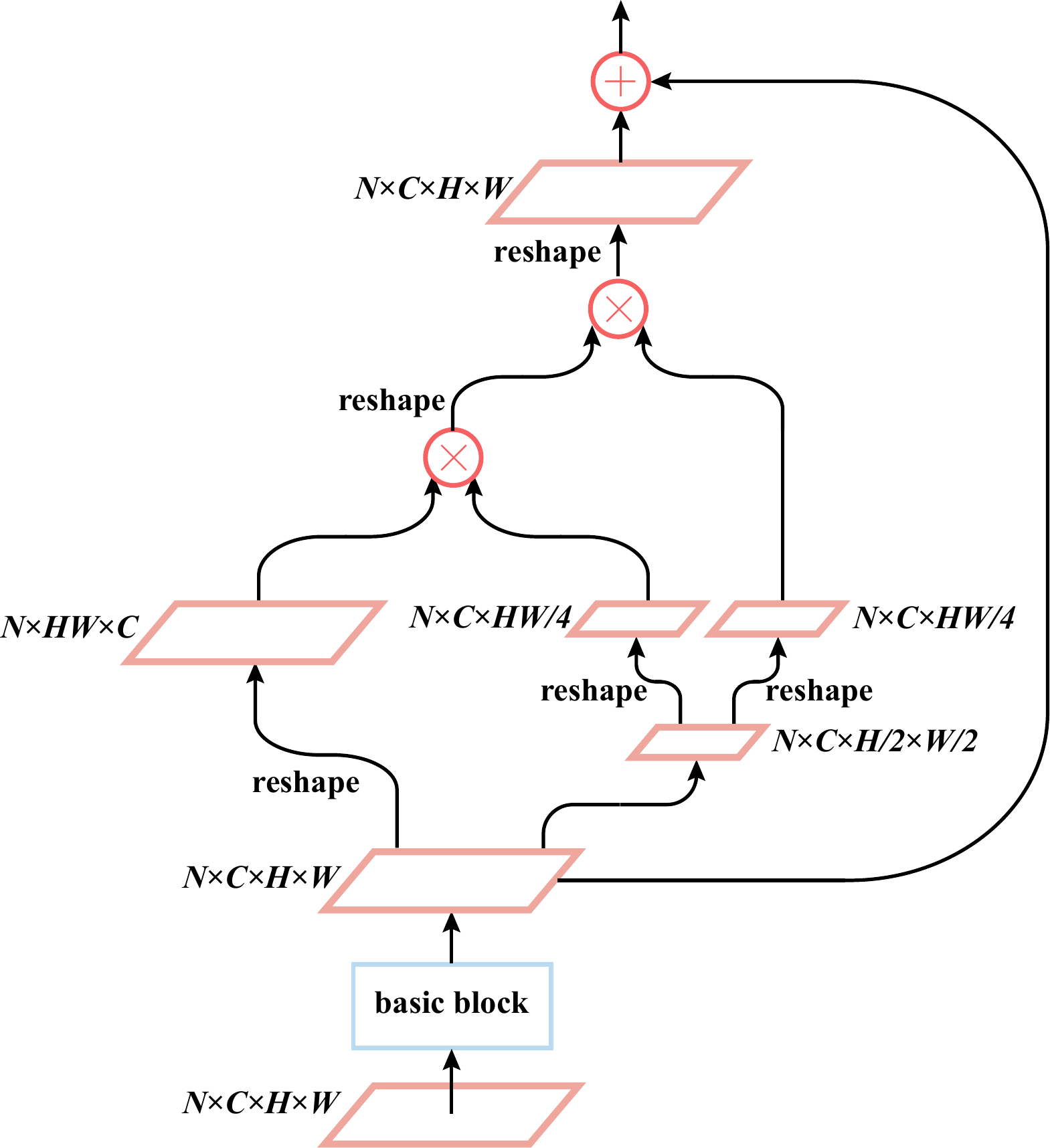}
\end{center}
\caption{Overview of the geometric attentional block in variational autoencoder. Basic block is a basic residual block in variational autoencoder. $\oplus$ means element-wise sum and $\otimes$ means matrix multiplication.}
\label{fig:avae}
\end{figure}

\noindent\textbf{Instance-level attentional block.} During the feature extracting and example generating, we plug geometric attentional blocks into VAE to constitute {\it attentional variational autoencoder}. In addition, we adopt channel-wise attentional blocks into the generator to model interdependencies between the channel to capture feature representation of faces named {\it attentional generator}. 
\begin{figure}[t]
\begin{center}
\includegraphics[width=0.9\linewidth]{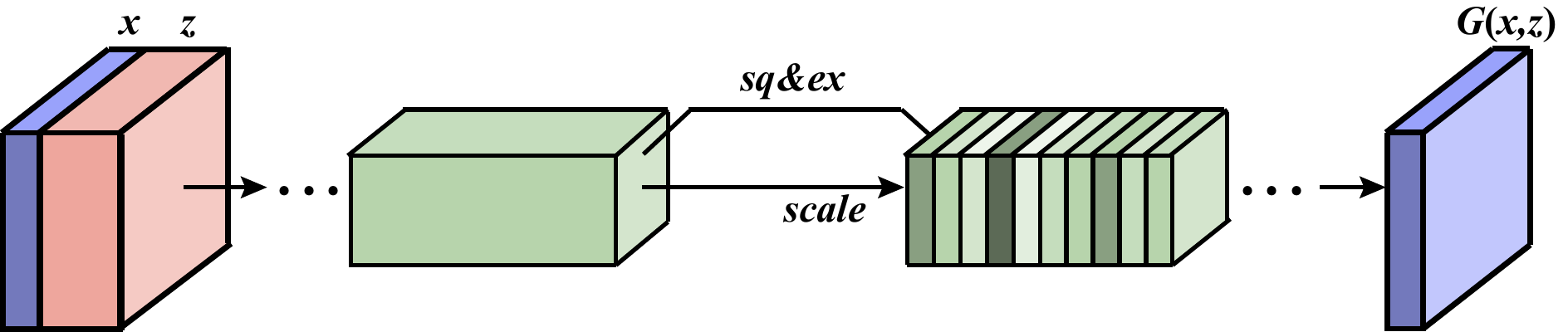}
\end{center}
\caption{Overview of a channel-wise attentional block in the generator. ``{\it sq\&ex}" is the squeeze operation (global average pooling) and the excitation operation (gating mechanism with a sigmoid activation). ``{\it scale}" is the operation to rescale the transformation output with activations after squeeze and excitation to get the channels with different weights of importance.} 
\label{fig:se}
\end{figure}

The overview of an attentional block in VAE is shown in Fig.~\ref{fig:avae}. VAE in our work is to learn the feature representation of the target person whose facial dependency is significant for capturing the latent code. It is related to the self-attention method which computes the response at one point in a sequence such facial feature by attending to all points. For this purpose, we introduce non-local block~\cite{non-local} to capture the facial dependency. For instance-level learning, we combine basic variational autoencoder residual block and non-local to propose {\it attentional VAE (AVAE)} in our $A^{3}GN$ in Fig.~\ref{fig:avae}. As shown in Fig.~\ref{fig:a3gn}, attentional VAE can encode the geometric information of target face and learn the facial dependency from different parts of human face effectively.

We concatenate the original face $x$ (3-dimension) with the latent code $z$ (7-dimension) as the input of attentional generator in Fig.~\ref{fig:se}. After two subsampling convolution layers in the generator, we introduce squeeze-and-excitation operations~\cite{senet} to emphasize informative features and suppress less useful ones in channel. SE operations propose to squeeze global spatial information into a channel descriptor by using global average pooling to generate channel-wise statistic. In excitation operation, a gating mechanism with a sigmoid activation is employed to capture channel-wise dependencies. Finally, we employ scaling to rescale the transformation output. Owing to squeeze-and-excitation, we can maintain informative features from the latent code more and suppress the useless information in channels which contributes to capturing feature representation of the target person. 

\section{Experiments}
\subsection{Evaluation}
In our work, we define a set of specific evaluation criteria to measure the effectiveness of the attacks:
\begin{itemize}
\vspace{-0.1cm}
\item
\noindent\textbf{Real accuracy \& Fake accuracy \& mAP.} It is defined as the percentage of adversarial examples which are successfully classified as the target person by target model. When cosine distance between examples and target faces is more than 0.45, we consider examples as target faces with a true predicting label. Real accuracy shows the percentage of original images which can be classified as the target person, which is usually 0\%, while fake accuracy shows the percentage of generated images which can be classified as the target person. mAP is the mean average precision with different thresholds in a range from 0 to 1 whose step is 0.01.
\item
\noindent\textbf{Similarity score.} Cosine distance between original faces/generated faces and target face faces is seen as a similarity score. Cosine distance is a significant metric in face recognition for verifying whether the two images belong to one person. In our results, we show the similarity scores before attack and after attack and improvement ($\Delta$) of similarity scores to exhibit the effectiveness of $A^{3}GN$ for attacks. Meanwhile, the similarity scores between the real image and the fake image exhibit the ability that the generated images can be recognized as their real identities by face recognition networks. The similarity scores between the real image and the fake image are less, the attack is more successful.
\item
\noindent\textbf{SSIM.} SSIM means the percentage of structural similarity index between original faces and generating faces higher than a threshold. SSIM is a quantization criterion to determine whether generating faces are perturbed slightly compared with original faces. In our work, we set 0.9 as the threshold to evaluate the quality of generated images compared to original images.
\end{itemize}

\subsection{Datasets}
The state-of-the-art face recognition networks are trained in CASIA-WebFace dataset and refined MS-Celeb-1M~\cite{arcface}. Meanwhile, our $A^{3}GN$ is also trained on CASIA-WebFace. And in the inference time, we perform $A^{3}GN$ on LFW by generating adversarial examples paired with target faces to verify whether they belong to one person.

\noindent\textbf{CASIA-WebFace.} CASIA-WebFace dataset~\cite{casia} is a web-collected dataset which has 494,414 face images belonging to 10,575 different individuals. In our experiments, we use aligned CASIA-WebFace which has images with size of 112$\times$112 after alignment.

\noindent\textbf{MS-Celeb-1M.} The original MS-Celeb-1M dataset~\cite{ms1m} contains about 100k identities with 10 million images. In ~\cite{arcface}, the noise of MS-Celeb-1M is decreased, and finally, refined MS-Celeb-1M contains 3.8M images of 85k unique identities. 

\noindent\textbf{LFW.} LFW dataset~\cite{lfw} contains 13, 233 web-collected images from 5749 different identities, with large variations in pose, expression and illuminations. In face verification, the verification accuracy is usually measured on 6000 face pairs. But in our work, we pair all the images in LFW with target face image.

\subsection{A$^{3}$GN with Attentional Block}
In this section, we do some experiments to verify the feasibility and effectiveness of attentional blocks.  We train $A^{3}GN$ on CASIA-WebFace and utilize it to generate adversarial examples on LFW to attack target model in the inference time. We employ ArcFace~\cite{arcface} which has an accuracy of 99.42\% on LFW as the target model in these experiments. 

\noindent\textbf{Network architecture.} We design $A^{3}GN$ based on cVAE-GAN. For the encoder, we use a classifier with 4 residual basic blocks for the latent code with 7 dimensions. Adapted from~\cite{stargan}~\cite{cyclegan}, generator in our work is composed of two convolution layers for downsampling, 6 residual blocks, and two convolution layers for upsampling. In the generator, we use instance normalization which are not used in discriminator. In our work, we have two discriminators. One is the target face recognition network for classifying whether the image patches belong to the target person or not called {\it instance discriminator} and another is PatchGAN discriminator~\cite{isola} for classifying whether the image patched are real or not called {\it normal discriminator}.

\noindent\textbf{Training details.} In the training process, the target person contains 7 different face images for capturing the latent code. All the input images are resized and cropped to 112$\times$112. Because our goal is to generate images for fooling the face recognition network, all the images should do the alignment similar to  the operation in face verification. We update generator once by $\mathcal{L}_{G}$ after five normal discriminator updates and one generator update by $\mathcal{L}_{cos}$ while the instance discriminator is fixed all the time. All the models are trained for 200000 iterations and use Adam~\cite{adam} with $\beta_1$ = 0.5 and $\beta_2$ = 0.999. The batch size is set to 32 in all experiments. We set the learning rate to 0.0001 for the first 100000 iterations and linearly decay the learning rate to 0 over the next 100000 iterations. 

\noindent\textbf{Quantitative evaluation.} We perform a quantitative analysis of the mAP, the difference of similarity score and SSIM on our baseline. All the results are calculated on average among 5 target faces to eliminate the occasionality. The performance of baseline is shown in Table~\ref{tab:baseline}. We design two groups of experiments with different conditions to verify the effectiveness of $A^{3}GN$. One is to encode image $A$ to attack image $A$. Another is to encode image $A$ to attack image $A'$. Neither of $A$ and $A'$ is in target images datasets for the latent code in the training process. In the experiment of baseline, we choose one target person randomly to test the performance. The threshold of cosine distance for fake accuracy is set to 0.45. The experiment in $A\to A$ can get higher accuracy than the experiment in $A\to A'$ because it can learn the feature representation of $A$ in the encoder for attacking $A$. As shown in the Table~\ref{tab:baseline}, our baseline can fool the target model generally. Most of them can be classified as the target person in the threshold of 0.45. SSIM is a criterion to evaluate the quality of generated images in some similar works. But we think it does not a strict criterion to evaluate the similarity between the generated images and the original images for human eyes.

We notice that the accuracy and mAP are not extremely high. For improving the performance, we consider introducing some attentional blocks to learn more feature representation of the target person. For capturing the facial dependencies of the target person, we introduce geometric attentional blocks, non-local blocks, into the encoder to improve the performance. And the performance is shown in Table~\ref{tab:nonlocal}. Obviously, each criterion gets improvement compared with the baseline. It is effective to capture the facial dependencies for encoding the latent code.

The geometric attention in encoder can capture the instance-level information effectively. We conjecture that introducing attentional blocks in the generator may also get better performance. During the process of generating images, the generator forces the fake images more similar to the original images which results in the loss of feature representation of the target person due to $\mathcal{L}_{rec}$. Thus, we consider introducing channel-wise attentional blocks into the generator to focus on the information of the latent code. The performance is shown in Table~\ref{tab:se}. It exceeds our expectations to outperform the experiment of geometric attentional blocks by 1.95\% of mAP in $A\to A$ condition. We conjecture that channel-wise attentional blocks maintain the instance information from the latent code primarily.

\begin{figure}[t]
\begin{center}
\includegraphics[width=0.95\linewidth]{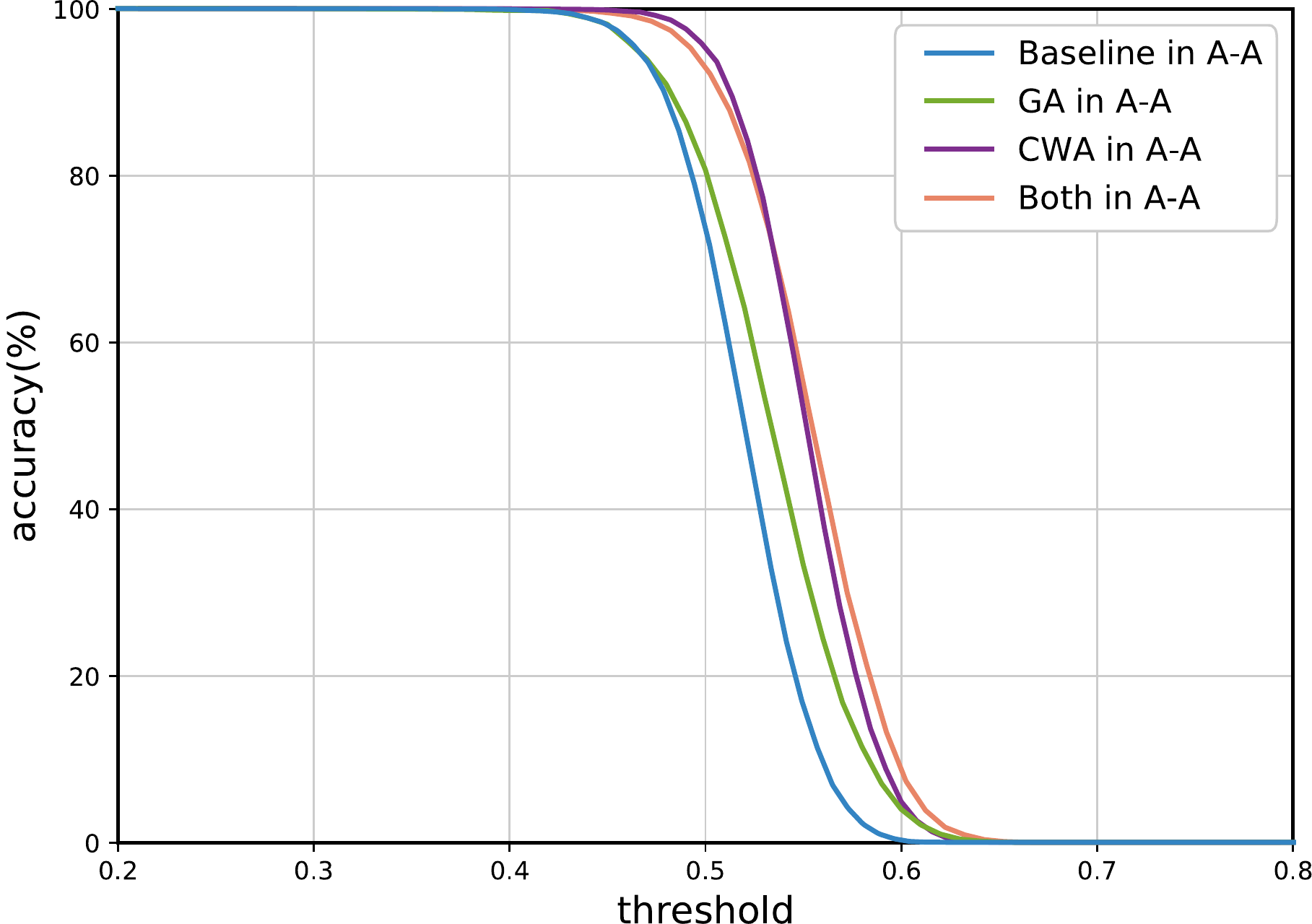}
\end{center}
\caption{Accuracy curve in different thresholds. The horizontal axis represents the different thresholds and the vertical axis represents the accuracy in different thresholds. GA means geometric attention. CWA means channel-wise attention.}
\label{fig:result}
\end{figure}

Following the two aforementioned experiments of attentional blocks, we combine geometric attention and channel-wise attention to improve the performance. The ablation study results are shown in Table~\ref{tab:attention}. And the curves of accuracies in $A\to A$ are shown in Fig.~\ref{fig:result}. As we can see, most of the generated images can get more than 0.4 of cosine distance which far surpasses the result between real images and the target image. $A^{3}GN$ can fool the face recognition network successfully.
\begin{table}\small
\setlength{\abovecaptionskip}{0pt}
\setlength{\belowcaptionskip}{-5pt}
\begin{center}
\resizebox{\linewidth}{!}{%
\begin{tabular}{c|c|c|c|c}
	\hline
   	 & Fake acc.(\%) &mAP(\%)&similarity score($\Delta$)&SSIM(\%) \\
	\hline
        \hline
         $A\to A$ & 97.52 &53.74& 0.506 & 3.466 \\
         $A\to A'$ & 93.82 &51.72& 0.490 & 3.57 \\
	\hline
\end{tabular}
}
\end{center}
\caption{Performance of $\textbf{baseline}$ with two conditions.}
\label{tab:baseline}
\end{table}
\begin{table}\small
\setlength{\abovecaptionskip}{0pt}
\setlength{\belowcaptionskip}{-5pt}
\begin{center}
\resizebox{\linewidth}{!}{%
\begin{tabular}{c|c|c|c|c}
	\hline
   	 & Fake acc.(\%) &mAP(\%)&similarity score($\Delta$)&SSIM(\%) \\
	\hline
        \hline
         $A\to A$ & 98.92 &54.62& 0.517 & 3.389 \\
         $A\to A'$ & 95.08 &52.53& 0.500 & 3.69 \\
	\hline
\end{tabular}
}
\end{center}
\caption{Performance of $\textbf{geometric attention}$ with two conditions.}
\label{tab:nonlocal}
\end{table}
\begin{table}\small
\setlength{\abovecaptionskip}{0pt}
\setlength{\belowcaptionskip}{-5pt}
\begin{center}
\resizebox{\linewidth}{!}{%
\begin{tabular}{c|c|c|c|c}
	\hline
   	 & Fake acc.(\%) &mAP(\%)&similarity score($\Delta$)&SSIM(\%) \\
	\hline
        \hline
         $A\to A$ & 99.67 &56.57& 0.533 & 4.609 \\
         $A\to A'$ & 98.96 &55.14& 0.523 & 4.48 \\
	\hline
\end{tabular}
}
\end{center}
\caption{Performance of $\textbf{channel-wise attention}$ with two conditions.}
\label{tab:se}
\end{table}
\begin{table*}[t]
\renewcommand\arraystretch{1.1}
\newcommand{\tabincell}[2]{\begin{tabular}{@{}#1@{}}#2\end{tabular}}
\begin{center}
\begin{tabular}{lp{3.5cm}<{\centering}|p{1.2cm}<{\centering}|p{1.2cm}<{\centering}|p{1cm}<{\centering}|p{1.5cm}<{\centering}|p{1.5cm}<{\centering}|p{1.5cm}<{\centering}|p{1cm}<{\centering}}
\hline
& \multirow{3}{*}{model} &  \multicolumn{3}{c|}{Attack acc. on LFW(\%)} & \multicolumn{3}{c|}{similarity score} &  \multirow{3}{*}{SSIM(\%)}\\
\cline{3-8}
 & & Real acc. & Fake acc. & mAP& Before $A^{3}GN$ & After $A^{3}GN$ & Between R\&F & \\
\hline
\hline
\multirow{4}{*}{\tabincell{c}{$A\to A$}}
& Baseline & 0.0 & 97.52 & 53.74 & 0.026 & $0.532_{(0.506)}$ &0.163 & 3.466\\
& Geometric attention & 0.0 & 98.12 & 54.62 & 0.026 & $0.543_{(0.517)}$& 0.161& 3.389\\
& Channel-wise attention & 0.0 & $\textbf{99.66}$ & 56.27 & 0.026 & $0.558_{(0.532)} $& $\textbf{0.146}$ & 4.609\\
& Both & 0.0 & 99.59 & $\textbf{56.28}$ & 0.026 & $\textbf{0.559}_{(0.533)}$ & 0.161& $\textbf{5.194}$\\
\hline
\hline
\multirow{4}{*}{\tabincell{c}{$A\to A'$}}
& Baseline & 0.0 & 93.82& 51.72 & 0.022 & ${0.512}_{(0.490)}$ & 0.163& 3.57\\
& Geometric attention & 0.0 & 95.08 & 52.53 & 0.022 & ${0.523}_{(0.500)}$ & 0.162 & 3.69\\
& Channel-wise attention & 0.0 & $\textbf{98.96}$ &54.99 & 0.022 & ${0.545}_{(0.523)}$ &$\textbf{0.146}$ & 4.48\\
& Both & 0.0 & 98.92 & $\textbf{55.14}$ & 0.022 & $\textbf{0.546}_{(0.525)}$ &0.160 & $\textbf{5.16}$\\
\hline
\end{tabular}
\end{center}
\caption{Ablation study of 4 different models for $A^{3}GN$. Baseline: Conditional GAN baseline. Geometric attention: Conditional GAN with geometric attentional blocks. Channel-wise attention: Conditional GAN with channel-wise attentional blocks. Both: Conditional GAN with geometric attentional and channel-wise attentional blocks. The threshold of cosine distance is set to 0.45.}
\label{tab:attention}
\end{table*}

\begin{figure}[t]
\begin{center}
\includegraphics[width=1.0\linewidth]{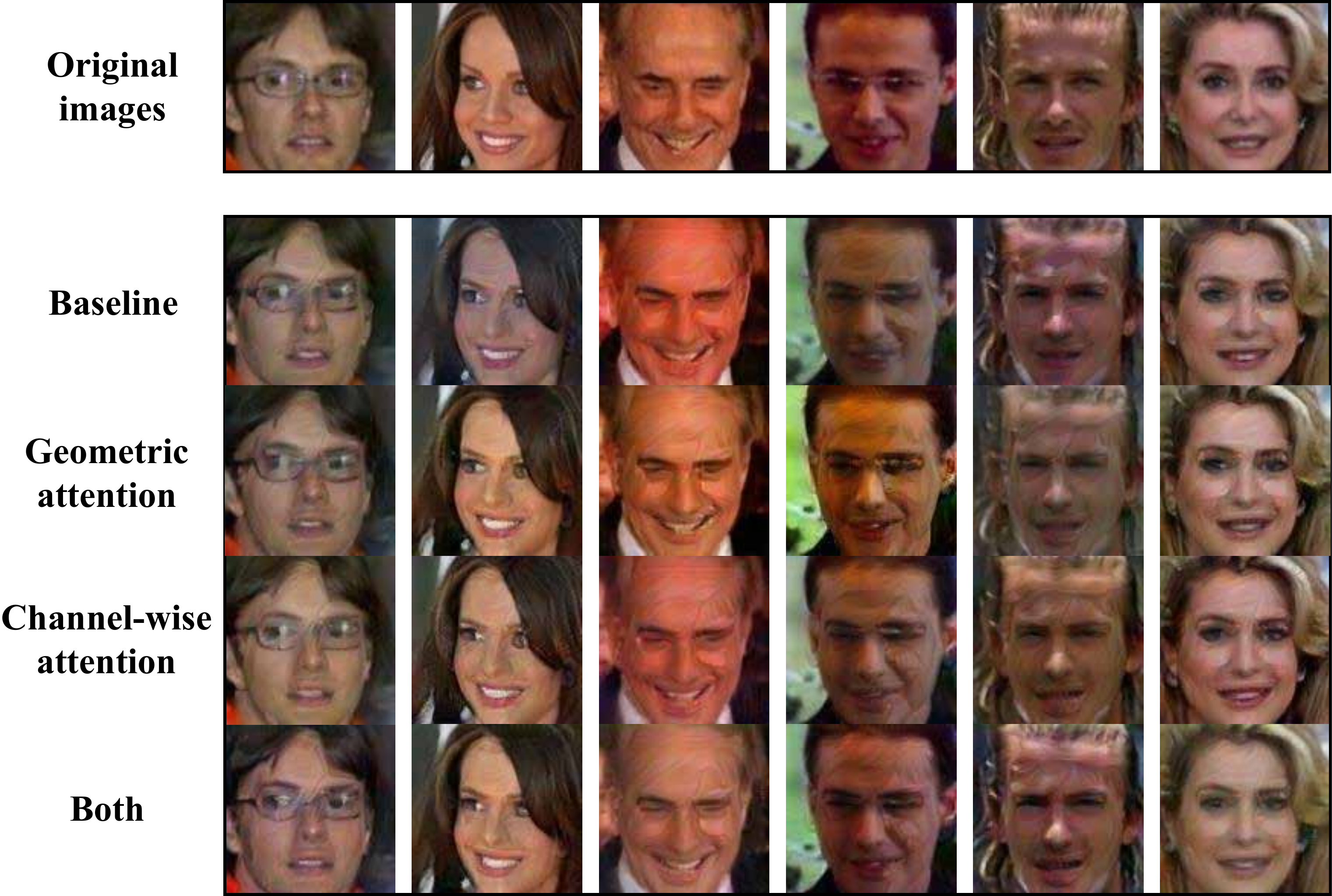}
\end{center}
\caption{Generated images by $A^{3}GN$ with 4 different models. The first row is the original images and the rest is the generated images by $A^{3}GN$ with 4 different models. The target person is Target A in Fig.~\ref{fig:5targets}.}
\label{fig:attention}
\end{figure}

\noindent\textbf{Qualitative evaluation.} In addition to the quantitative evaluation, we exhibit the effectiveness of 4 different models by showing the qualitative comparison results in Fig.~\ref{fig:attention}. All the generated images in Fig.~\ref{fig:attention} can be classified as the target person in the threshold of 0.45 and they are similar to the original images just with quasi-imperceptible perturbation. We observe that our model can provide a higher visual quality of attack results on LFW even in baseline. However, the generated images are similar to the target image in physical likeness slightly such as the nose and eyes. We conjecture that it is because that the generator hammers at making the cosine distance between generated images and target image higher. It shows that face recognition network recognizes people by focusing on their noses and eyes more, and the contours of their faces and mouths less.

\begin{figure*}
\begin{center}
\includegraphics[width=0.95\linewidth]{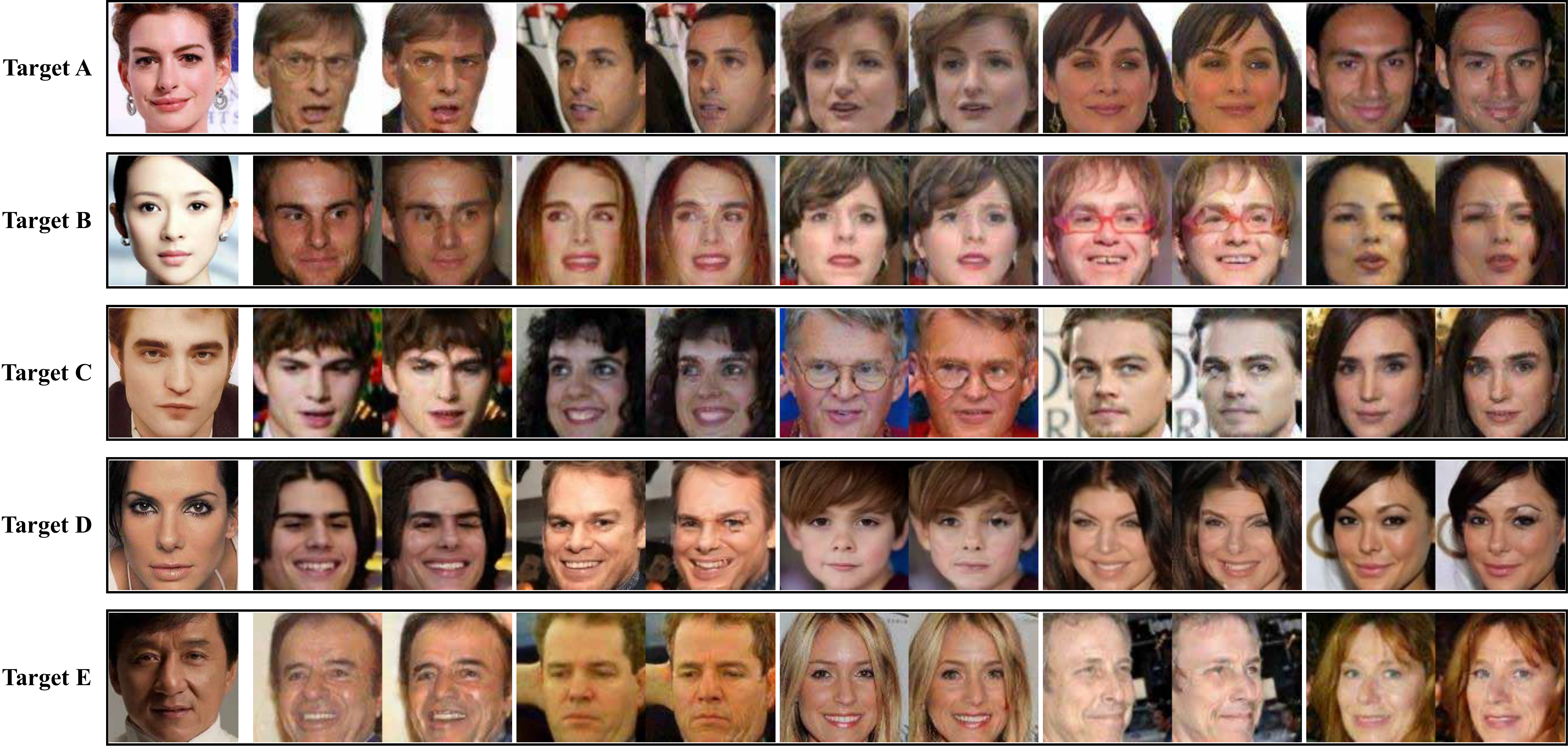}
\end{center}
\caption{Generated images by $A^{3}GN$ for 5 target faces. The left is original image and the right is generated image.}
\label{fig:5targets}
\end{figure*}

Furthermore, we choose 5 different target face images to exhibit the results of attacks in Fig.~\ref{fig:5targets}. Most of the generated images are prone to the target face image slightly. It would seem that most face recognition network focuses on recognizing people by their facial feature and a slight change on the facial feature can fool the face recognition network to recognize as another person which is imperceptible for observers. Meanwhile, a mask learned from the target person can also fool the network.

\noindent\textbf{White-box attack.} In white-box scenario, we choose 3 different state-of-the-art face recognition networks to verify the feasibility of our model $A^{3}GN$ in different target networks. The performance on different target models in white-box scenario is shown in Table~\ref{tab:white-box}. In white-box scenario, the parameters, architectures and the feature space of target models are obtained in the training process. Thus, the generator can generate images directionally. The metrics of evaluation in this section are mAP, the difference of similarity score, the similarity score between real images and fake images and SSIM. And in Table~\ref{tab:white-box}, we show the accuracy of the target network on LFW in face verification. mAP and the difference of similarity score indicate the ability to fool the networks to recognize as the target person and the similarity score between real images and fake images can indicate the ability to fool the networks to be mistaken. All of them can prove that our model $A^{3}GN$ can be applied to fool different state-of-the-art networks.

In the experiment on attacking ResNet with softmax and Sphereface, mAPs are lower than that in the experiment on attacking ArcFace. But the attack on Sphereface is more effective in reducing the similarity score between real images and fake images. And we conjecture that different training data and different accuracy on LFW result in the different performance on generated images. For ResNet with softmax and Sphereface, the training data is CASIA-WebFace. But for ArcFace, the training data is MS-Celeb-1M. Different training data bring about different feature representation. 
\begin{table}\small
\setlength{\abovecaptionskip}{0pt}
\setlength{\belowcaptionskip}{-5pt}
\begin{center}
\resizebox{\linewidth}{!}{%
\begin{tabular}{c|c|c|c|c|c}
	\hline
   	 \multirow{2}{*}{target model}&\multirow{2}{*}{acc. on LFW(\%)}&\multirow{2}{*}{mAP(\%)}&\multicolumn{2}{c|}{similarity score} &\multirow{2}{*}{SSIM(\%)} \\
	 \cline{4-5}
	 & & & $\Delta$ & R\&F & \\
	\hline
        \hline
        Softmax & 97.02 & 45.32 & 0.421 & 0.232& 5.521 \\
	Sphereface~\cite{cosineface} & 99.20 & 49.04 & 0.478 &0.090 &6.440 \\
	ArcFace~\cite{arcface} & 99.42 & 56.26 & 0.530 & 0.161 & 5.025 \\
	\hline
\end{tabular}
}
\end{center}
\caption{$A^{3}GN$ performance on different target models in white-box scenario.}
\label{tab:white-box}
\end{table}

\noindent\textbf{Black-box attack.} In this section, we explore whether fooling one face recognition network leads to successful fooling other networks. In black-box scenario,  the parameters, architectures and the feature space of target models are not obtained in the training process. The instance discriminator in black-box scenario is only ArcFace~\cite{arcface} in this experiment. And we have no access of target networks, ResNet~\cite{resnet} with softmax, Sphereface~\cite{sphereface} and MobileFaceNet~\cite{mobileface} in the training process. In the inference time, we just obtain the feature map of images from the last layers to test the performance. The performance on different target networks in black-box scenario is shown in Table~\ref{tab:black-box}. Obviously, each result in Table~\ref{tab:black-box} is lower than that in Table~\ref{tab:white-box}. But we also observe that the generated images can disturb the target networks slightly. Black-box attack will be a future work to explore and study.
\begin{table}\small
\setlength{\abovecaptionskip}{0pt}
\setlength{\belowcaptionskip}{-5pt}
\begin{center}
\resizebox{\linewidth}{!}{%
\begin{tabular}{c|c|c|c|c}
	\hline
   	 \multirow{2}{*}{target model}&\multirow{2}{*}{acc. on LFW(\%)}&\multirow{2}{*}{mAP(\%)}&\multicolumn{2}{c}{similarity score} \\
	 \cline{4-5}
	 & & & $\Delta$ & R\&F\\
	\hline
        \hline
        Softmax & 97.02 & 17.70& 0.082 & 0.58\\
	Sphereface~\cite{sphereface} & 99.20& 12.32 & 0.110 & 0.59\\
	MobileFaceNet~\cite{mobileface} & 99.18& 9.07 & 0.297 & 0.41\\
	\hline
\end{tabular}
}
\end{center}
\caption{$A^{3}GN$ performance on different target models in black-box scenario.}
\label{tab:black-box}
\end{table}

\noindent\textbf{Comparison with previous works.} We compare our $A^{3}GN$  with previous attack models in face recognition on CASIA-WebFace dataset. Because they focus on fool the classifier to a false label, we compare our performance on this way in Table~\ref{tab:compare}. If the cosine distance between the original image and the generated image is lower than 0.45, it is seen as a success for an attack. As we can see, the success rate of fool the face recognition network to a false label for $A^{3}GN$ is 99.94\%. It almost fools the network totally. Though it is 0.02\% lower than GFLM, $A^{3}GN$ can force the target model to recognize as the target person well.

\begin{table}\small
\setlength{\abovecaptionskip}{0pt}
\setlength{\belowcaptionskip}{-5pt}
\setlength{\tabcolsep}{7mm}
\begin{center}
\resizebox{\linewidth}{!}{%
\begin{tabular}{c|c|c}
	\hline
   	 model& SR(\%) & Attack acc. on CASIA(\%)\\
	\hline
        \hline
        stAdv~\cite{stadv} & 99.18 & -\\
	GFLM~\cite{gflm} & 99.96& - \\
	$A^{3}GN$ & 99.94& 98.23 \\
	\hline
\end{tabular}
}
\end{center}
\caption{Comparison with other attack models in face recognition. `SR' means the success rate of fooling the network to a false label. `Attack acc. on CASIA' means the accuracy of fooling the network to a target label.}
\label{tab:compare}
\end{table}

\section{Conclusion}
Face recognition is a compelling task in deep learning. It is necessary to learn how face recognition networks are subject to attacks. In this paper, we focus on a novel way of attacking target models by fooling them to a specific label. For this purpose, we present $A^{3}GN$ to generate adversarial examples similar to the original images but which can be classified as the target person. To learn the feature representation of target images, we introduce geometric attention and channel-wise attention into $A^{3}GN$ to get good performance. Finally, we show the results of experiments on different target faces, white-box attack, and black-box attack. However, our model is limited to attacking one target person. It will be a future work that one model can attack different target faces.
{\small
\bibliographystyle{ieee}
\bibliography{egbib}
}

\end{document}